\documentclass[a4paper, 10 pt, journal]{ieeeconf}  

\IEEEoverridecommandlockouts                              

\usepackage{graphics} 
\usepackage{epsfig} 
\usepackage{amsmath}
\usepackage{color}
\usepackage{url}
\usepackage{subfigure}
\usepackage[ruled,vlined,linesnumbered]{algorithm2e} 
\usepackage[table]{xcolor}
\usepackage{lipsum}

\newcommand{\zhi}[1]{\textcolor{black}{#1}}

\newcommand{\new}[1]{\textcolor{black}{#1}}

\title{\LARGE \bf
Online Context Learning for Socially Compliant Navigation
}

\author{Iaroslav Okunevich$^1$, Alexandre Lombard$^1$, Tomas Krajnik$^2$, Yassine Ruichek$^1$, Zhi Yan$^{1,3*}$
  \thanks{This work was supported by the Bourgogne-Franche-Comt\'e regional research project LOST-CoRoNa, the CZ MSMT (No. 8J23FR023) and PHC Barrande (No. 49275QM) project 3L4AR, and the Project Robotics and Advanced Industrial Production [no. CZ.02.01.01/00/22\_008/0004590].}
  \thanks{This work involved human subjects or animals in its research. Approval of all ethical and experimental procedures and protocols was granted by the ethics committee of the CIAD laboratory.}
  \thanks{$^1$UTBM, CIAD UMR 7533, F-90010 Belfort, France. {\tt\small firstname.lastname@utbm.fr}}%
  \thanks{$^2$Faculty of Electrical Engineering, Czech Technical University in Prague, Prague, Czechia.
  {\tt\small krajnt1@fel.cvut.cz}}%
  \thanks{$^3$U2IS, ENSTA, Institut Polytechnique de Paris, France.}%
  \thanks{$^*$Corresponding Author.}%
}

\begin{document}

\maketitle
\thispagestyle{plain}
\pagestyle{plain}

\begin{abstract}
\zhi{Robot social navigation needs to adapt to different human factors and environmental contexts.
However, since these factors and contexts are difficult to predict and cannot be exhaustively enumerated, traditional learning-based methods have difficulty in ensuring the social attributes of robots in long-term and cross-environment deployments.
This letter introduces an online context learning method that aims to empower robots to adapt to new social environments online.
The proposed method adopts a two-layer structure.
The bottom layer is built using a deep reinforcement learning-based method to ensure the output of basic robot navigation commands.
The upper layer is implemented using an online robot learning-based method to socialize the control commands suggested by the bottom layer.
Experiments using a community-wide simulator show that our method outperforms the state-of-the-art ones.
Experimental results in the most challenging scenarios show that our method improves the performance of the state-of-the-art by 8\%.}
The source code of the proposed method, the data used, and the tools for the per-training step are publicly available at \url{https://github.com/Nedzhaken/SOCSARL-OL}.
\end{abstract}

\section{Introduction}

Mobile robots for work in human-presence, such as supermarket cleaning~\cite{zhimon20jist} or hospital disinfection~\cite{perminov2021ultrabot} robots are a crucial research domain. They need to take into account human behavior and their social attributes in their task and motion planning~\cite{vintr20iros}.
In these applications, robots must achieve high-quality human-robot interaction (HRI) to ensure that human comfort is not compromised~\cite{okunevich2023human}.

Socially compliant robot navigation is a software solution to achieve effective HRI.
The idea is to apply human social rules to HRI while avoiding collisions with people.
The rules applied can be based on environmental~\cite{shahrezaie2022advancing} or social~\cite{benedictis2023dichotomic} context.
The social compliance that a robot should demonstrate when navigating can be achieved through special reward functions in deep reinforcement learning (DRL) algorithms~\cite{chen2017socially}, attention scores of people around the robot~\cite{chen2019crowd}, spatiotemporal representations of people around~\cite{yang2023st}, and so forth~\cite{okunevich2023human}.
Unlike humans, who adapt to changing social conditions, these methods rely on predefined rules.

Key challenges in robot social navigation are planning the robot's task and motion in a socially compliant manner, and evaluating different approaches.
In our previous work, we focused on standardizable evaluation procedures for social navigation~\cite{okunevich2023human}.
In this letter, we focus on the behavior of robots.
Our research goal is to ensure the social efficiency and navigation robustness of mobile robots when deployed in new social scenarios.
An online robot learning (ORL) approach is proposed to achieve this goal, as it is able to train knowledge (discriminative) models spontaneously and autonomously in new contexts without manual intervention.

\begin{figure}
  \centering
  \includegraphics[width=\columnwidth]{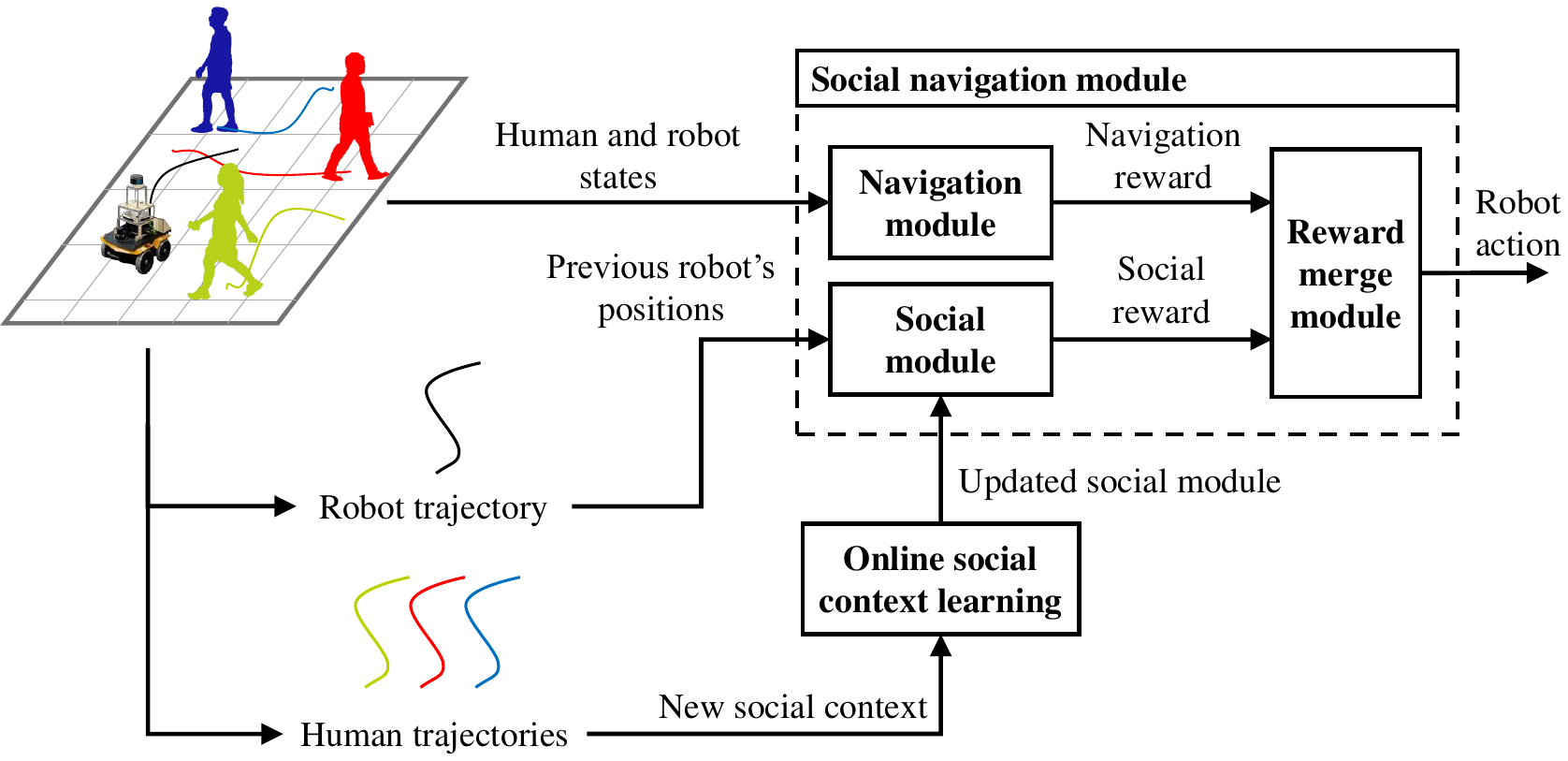}
  \caption{Conceptual diagram of our proposed approach. The navigation module selects the robot's next action. The social module adds social value to the proposed action. The online social context learning method updates the social module to adapt it to the new social environment, which is represented by human trajectories.}
  \label{fig:concept}
\end{figure}

\zhi{A conceptual diagram of our method is shown in
Fig.~\ref{fig:concept}.
Specifically, we first utilize DRL as a training algorithm for our foundation model to control the robot motion, as it is proven effective for robot navigation in crowds~\cite{chen2019crowd}.
We then build a ``social module'' to evaluate the social efficiency of the robot trajectory and adjust the robot motion to achieve socially compliant behavior by intervening in the foundation model.
The module aims to extract the hidden social rules of human movement and is initially trained with human social trajectory data.
However, it is likely that the social context of these training data does not correspond to the new one in which the robot is deployed.
Therefore, the module is adapted to new contexts by being updated on-the-fly with on-site data while the robot is operating.
}

\zhi{The contribution of this letter is threefold.
\begin{itemize}
  \item We propose to combine DRL with ORL to enable mobile robots to adapt to new social contexts efficiently and autonomously.
  Its specific implementation includes building a social module that can be updated online and promptly intervening in the robot's DRL-based navigation system to ensure the robot's social efficiency after switching contexts, without human intervention.
  \item The social module is initially trained on the state-of-the-art (SOTA) contextually-rich THOR-Magni dataset~\cite{schreiter2024th}, which contains social trajectories of humans in shared spaces with a robot. We supplement this dataset with non-social trajectory data.
  \item We evaluate various social navigation methods, both ours and others, in new social contexts that are different from the one in which these methods were originally trained. We also complicate the experimental conditions to understand the robustness of the methods.
\end{itemize}
}

\section{Related Work}
\label{sec:related_work}

\subsection{Socially Compliant Navigation}

\zhi{A mainstream approach to enabling robots to exhibit social behaviors during navigation is based on learning, represented by imitation learning and DRL.
For the former, the simulated data can be the robot behavior controlled by humans in a real environment~\cite{karnan2022socially} or generated in a simulated environment~\cite{tai2018socially}.
It is worth pointing out that the human agents in~\cite{tai2018socially} are controlled by the social force (SF) model.
Multi-agent trajectory prediction is another application of imitation learning that can be applied to copy human behavior~\cite{bhattacharyya2019simulating}.
The limitations of existing imitation learning methods still lie in their reliance on expert demonstrations and their difficulty in adapting to new social environments.}

\zhi{DRL learns robot control policies through reward functions and the agent's state information.
The research axis most relevant to our work starts with the Collision Avoidance with DRL (CADRL) method proposed by Chen \emph{et al.}~\cite{chen2017decentralized}.
This method uses a neural network to parameterize the value function.
Subsequently, a socially-aware CADRL, named SA-CADRL~\cite{chen2017socially}, was proposed with a modified reward function that enables the robot to obey basic social rules, such as the right- or left-hand rule, when navigating.
One limitation of CADRL-based approaches is that the input dimension of the neural network used is fixed, which prevents the robot from taking advantage of information about more people around it.}

\zhi{Everett ~\emph{et al.}~\cite{everett2018motion} broke this limitation and utilized the data of all people around the robot by sending their states sequentially to a long short-term memory (LSTM) network.
One limitation of this approach is that the priority of the LSTM input is only inversely proportional to the distance between the human and the robot, i.e., the input with a shorter distance is prioritized, while their relative speed is ignored.
As a result, the impact of a human approaching the robot at high speed from a distance may be considered less important than a stationary person near the robot.}

Chen \emph{et al.}~\cite{chen2019crowd} proposed a solution to this problem. The idea comprises several steps. Since each person influences other people, the human-human pairwise interaction is modeled by a cost map. An embedding vector is then calculated for each human, based on their state and the cost map. The next step is calculating the attention score for each individual based on the individual embedding vector and the mean embedding vector of all detected persons. The final representation of the people around the robot is a linear combination of the individual attention score and the pairwise interaction vector of each person. This crowd representation facilitates efficient crowd navigation, an essential element of social navigation. Nevertheless, this method does not incorporate social rules, except for the requirement of social distance in the reward function.

The preceding methods, except the LSTM method~\cite{everett2018motion}, are predicated solely on the spatial dimension of HRI. Liu \emph{et al.}~\cite{liu2021decentralized} proposed a Decentralized Structural Recurrent Neural Network (DSRNN) to leverage the spatial and temporal relationships for crowd navigation. Yang \emph{et al.}~\cite{yang2023st} suggested applying a spatial-temporal state transformer for more effective crowd navigation. While the spatial-temporal transformer structure appears to be a more effective baseline for the social navigation method, the evaluation below will demonstrate that this method is less effective in crowd navigation compared with the attention mechanism~\cite{chen2019crowd}.

\subsection{Online Context Learning}

\zhi{The aforementioned methods are all based on offline trained models and are therefore challenged by new social contexts.
These contexts are diverse and even humans understand them differently~\cite{singamaneni2021human}.
There are only a few works related to ours.}
Shahrezaie \emph{et al.}~\cite{shahrezaie2022advancing} proposed an online context robot navigation. Contexts and corresponding navigation rules are predefined based on the interviews. The content detection system allows the robot to change navigation online, depending on the environment.
Lui \emph{et al.}~\cite{liu2021lifelong} proposed Lifelong Learning for Navigation (LLfN), which includes an additional module that corrects the robot's behavior in complex contexts and updates online to learn new contexts.
Although this work focuses on the problem of forgetting navigation experience and does not include social contexts, the structure of a main navigation method and an extra correction module is similar to our proposal.

\subsection{Discussion}

\zhi{According to our investigation, existing methods still struggle to meet the performance requirements of robots for social navigation in changing environments or across environments.
It is an intuitive idea to directly perform online context updates on the social navigation model.
However, taking the SOTA method SARL as an example, there are two obvious challenges.
First, an update is difficult to complete in a short time, which goes against the spirit of ``fast learning and immediate deployment'' of online robot learning~\cite{yz23aaai}.
Second, the update of the model also has a clear demand for computing resources, which poses a challenge to deploying the model to the edge~\cite{okunevich2024open}.}

\zhi{Therefore, considering the current algorithm and hardware development, it is a good choice to modularize the social context and only update the social module online.
Additional benefits of doing so include, first, the developed social module can also be easily integrated into other navigation systems.
Second, this solution is more in line with the requirements of the robustness of the robot system, that is, the social module will not harm the output of the robot navigation module, but only serves as an auxiliary.
Finally, it has better interpretability compared to the end-to-end methods~\cite{chen2017socially, chen2019crowd, yang2023st, everett2018motion}.}

\section{Method}

\begin{figure*}[t]
  \centering
  \includegraphics[width=\textwidth]{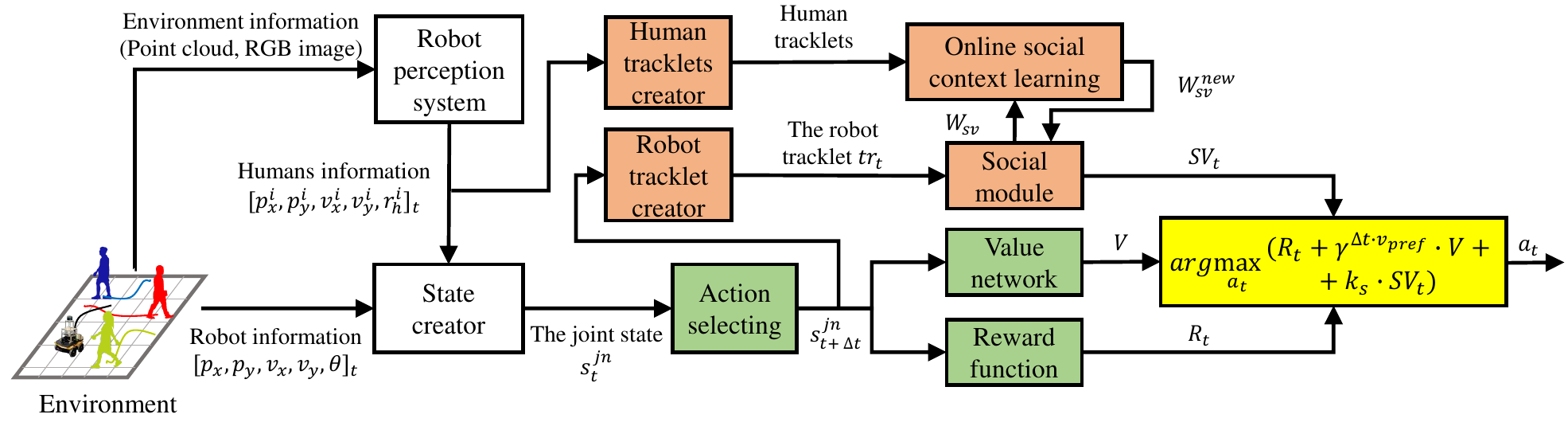}
  \caption{The overview diagram of our method. Based on human and robot states, the control algorithm identifies the optimal action from the action space. The human information is applied to update the social module. The 'human' and 'robot' tracklets creator blocks are identical. The orange and green blocks are the ORL-based social module and the DRL robot navigation elements respectively. The yellow block is the combination of ORL and DRL results.}
  \label{fig:overview_diagram}
\end{figure*}

We propose a novel architecture for robot social navigation that consists of two layers.
The underlying layer is a DRL-based robot navigation method due to its ability to handle complex navigation problems.
In addition, at this layer we extend considerations of safe distance to social distance.
SARL is therefore used in our concrete implementation due to its demonstrated SOTA performance.
The upper layer is a novel ORL-based social module that learns social context from human trajectory data and socializes the robot navigation control commands output by the bottom layer. In order to adapt to different social environments, the module is updated online based on new social contexts. The overview diagram is shown in Fig.~\ref{fig:overview_diagram}.
In the remainder of this section, we first formulate the target problems, then propose an improvement to the original SARL, and finally detail our online learnable social module.

\subsection{Problem Formulation}
\label{sec:Problem Formulation}

\zhi{Corresponding to reinforcement learning (RL), robot navigation can be formalized as a Markov decision process (MDP).
The objective is to determine the optimal strategy for robot control through interaction with the environment.
In social navigation, in addition to the environment, the robot should also consider the social attributes of humans:
\begin{equation}
\begin{split}
J(\pi) = & \sum_{s \in \mathcal{S}} \rho(s) \sum_{a \in \mathcal{A}} \pi^{*}(a|s) \cdot \\
& \sum_{s' \in \mathcal{S}} P(s'|s, a) [R(s, a) + \gamma H(s, a)],
\end{split}
\end{equation}
where $\sum_{s \in \mathcal{S}} \rho(s)$ represents all possible states $s$ in the environment, weighted by the initial state distribution $\rho(s)$;
$\sum_{a \in \mathcal{A}} \pi(a|s)$ indicates all possible actions $a$ available in a state $s$, weighted by the optimal policy's probability of taking that action $\pi^{*}(a|s)$;
$\sum_{s' \in \mathcal{S}} P(s'|s, a)$ represents all possible next states $s'$ that the robot could transition to after taking action $a$ in state $s$, weighted by the transition probability $P(s'|s, a)$;
$R(s, a)$ is the immediate reward received for taking action $a$ in state $s$;
$\gamma$ is the discount factor;
and $H(s, a)$ is the human-related cost function evaluated at taking action $a$ in state $s$.
$H(s, a)$ captures the impact of the environment's state on the robot's objective, considering human presence or behavior, which can be implemented based on hand-crafted features or learned models.
A specific implementation of this function is detailed in Section~\ref{sec:orl}.}

\zhi{On the other hand, since social environments are inherently changing, robots are expected to be able to update their policies online in order to adapt to the changes:
\begin{equation}
\label{eq.:f_d}
\pi(a | s, t+1) = f_d(\pi_{old}(a | s, t), R_t(s, a), t+1),
\end{equation}
where $\pi(a | s, t+1)$ represents the policy at time $t+1$;
and $\pi_{old}(a | s, t)$ represents the policy at time $t$.
Building an effective policy adaptation function $f_d$ is the key.}

\zhi{More specifically,} MDP includes the next important components: the joint state $s^{jn}_{t}$ at time $t$, action of agent $a_{t}$ at time t, reward $R_t(s^{jn}_{t}, a_{t})$ at time t, which the agent receive for its action and transition probability $P_t(s^{jn}_{t}, a_{t}, s^{jn}_{t + \Delta t})$.
The joint state at time \emph{t} equals a combination of robot-human states at time \emph{t} $s^{jn}_{t} = [rh^1_t, ... , rh^n_t]$, where $rh^n_t$ is the joint robot-human state with relative information between \emph{n}th human and the robot.
The robot-human state is a combination of robot and \emph{i}th human state around the robot $rh^i_t = [s^{r}_{t}, h^i_t]$. These states are defined as:
\begin{equation}
\label{eq:state}
  s^r_t = [p_x, p_y, v_x, v_y, r_r, g_x, g_y, \theta, v_{pref}]_t,
\end{equation}
\begin{equation}
  h^i_t = [p^i_x, p^i_y, v^i_x, v^i_y, r^i_h]_t,
\end{equation}
where $p_x$, $p_y$ and $p^i_x$, $p^i_y$ represent the robot and \emph{i}th human positions, $v_x$, $v_y$ and $v^i_x$, $v^i_y$ are the robot and \emph{i}th human velocities, $r_r$ and $r^i_h$ are the robot and \emph{i}th human radius. $g_x$, $g_y$ are the goal positions, $\theta$ is the robot orientation and $v_{pref}$ is the preferred robot speed. Position, speed, and radius are observable parameters and are available to other agents. The robot-human states are updated to follow the robot-centric parameterisation, where the x-axis points to the robot's goal and the robot's position is the origin. The final shape of the \emph{i}th robot-human state, which forms the input for the value neural network, is:
\begin{equation}
\begin{split}
{rh}^i_t = [d_g, v_{pref}, \theta, r_r, v_x, v_y, \\
p^i_x, p^i_y, v^i_x, v^i_y, r^i_h, d_i, r_{sum}]_t,
\end{split}
\end{equation}
where $d_g$ is the distance from the robot to the goal, $d_i$ is the distance to the \emph{i}th human, and $r_{sum}$ is the sum of robot and \emph{i}th human radius.

\subsection{SARL}
\label{sec:drl}

In SARL's value-based reinforcement learning framework, the goal is to discover the optimal navigation policy, denoted as $\pi^*(s^{jn}_t)$, to achieve the highest value $V^*(s^{jn}_t)$ for the joined state $s^{jn}_t$ at time $t$:
\begin{equation}
  V^*(s^{jn}_t) = \sum_{i=t}^{T} \gamma^{i\cdot v_{pref}}\cdot R_t(s^{jn}_{t}, \pi^*(s^{jn}_t)),
\end{equation}
where $\gamma \in [0,1)$ is the discount factor, $v_{pref}$ is used as a normalization parameter of $\gamma$ for numerical reasons~\cite{chen2017decentralized}, and $T$ is the time of the final state.
A constant velocity model is applied to estimate the subsequent states of humans over a short temporal interval $\Delta t$ and to approximate the next joint state $s^{jn}_{t + \Delta t}$ based on $s^{jn}_{t}$ and $a_t$, then the optimal strategy can be formulated as:
\begin{equation}
\label{eq:SARL}
  \pi^*(s^{jn}_t) = \arg \max_{a_t}[R_t(s^{jn}_{t}, a_{t}) + \gamma^{\Delta t\cdot v_{pref}}\cdot V(s^{jn}_{t + \Delta t})],
\end{equation}
where the value network is trained by the temporal-difference method with experience replay~\cite{chen2017socially, chen2019crowd, yang2023st}.

The speed and angular direction define the robot's action space. The holonomic kinematics have modeled the movements, enabling the agent to move in any direction with any acceleration. The action space contains 80 actions. The five linear velocities are distributed between $0$ to $v_{pref}$ and the angular directions are spaced from $0$ to $2\pi$. To select the best action, the sum of the reward and value functions is calculated for each action from the action space.

In original SARL, the reward is equal to 1 if the robot reaches the target position.
This reward is changed to $d_{plan}/d_{real}$ in ours, where $d_{plan}$ is the euclidean distance from its current position to the target one before the robot starts moving, and $d_{real}$ is the actual distance traveled by the robot after it reaches the specified location.
The interpretation of this change is that the robot should minimize unnecessary local path adjustments while maintaining socially compliant navigation.
Thus the full reward function becomes:
\begin{equation}
\label{eq:reward}
  R_t(s^{jn}_{t}, a_{t}) = \begin{cases}
      -0.25 & \text{if $d_{min} < 0$}\\
      d_{plan}/d_{real} & \text{else if $d_g = 0$}\\
      -0.1 + d_{min}/2 & \text{else if $d_{min} < d_c$}\\
      0 & \text{otherwise}
    \end{cases},
\end{equation}
\zhi{where $d_{min}$ is the distance between the robot and the nearest person, and $d_{c}$ is the comfortable social distance.
$d_{c}$ should be adjusted according to the spatio-temporal characteristics.
For example, it can be higher in large malls or during epidemics, and lower in small stores or during normal times.}

\subsection{Social Module}
\label{sec:orl}

\begin{figure}
\begin{center}
\subfigure[\label{fig:exam_trakclet}The robot tracklet $tr_t$. The blue dots are $u$ previous states of the robot, the yellow dot is its current state, the red dot is the robot's state after action $a_t$, and the green dots are $f$ future prediction states.]{
\includegraphics[width=1.00\linewidth]{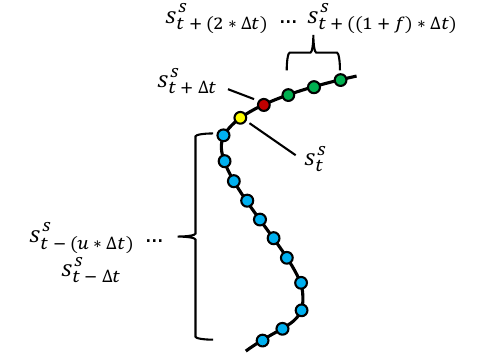}}
\subfigure[\label{fig:SVNN_example}The structure of social value network with GRU and four fully-connected (FC) layers. The input is the robot tracklets. The numbers at the bottom are the input dimensions. Social value (SV) is the output of this network and can be 1 for social tracklet and 0 for non-social tracklet.]{
\includegraphics[width=1.00\linewidth]{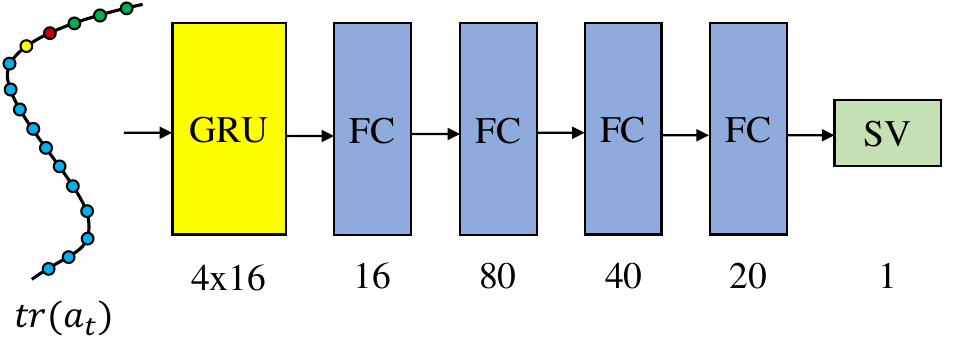}}
\caption{A example of robot tracklet and the proposed social value network.}
\label{2exp_result}
\end{center}
\end{figure}

\zhi{The task of the social module is to implement social modifications to the SARL output.
Intuitively, if the action $a_{t}$ suggested by SARL is considered to be social, its corresponding value should be increased so that it is selected by the optimal policy $\pi^*(s^{ jn}_t)$.
Otherwise, its corresponding value should be lowered so that it is discarded by the policy $\pi^*(s^{jn}_t)$.
Based on the dual consideration of the effectiveness of small-batch online learning~\cite{yz19auro} and tracklet-based behavioral analysis~\cite{serhan17tcsvt},
we employ a tracklet-based sociality assessment method, which is visualized in Fig.~\ref{fig:exam_trakclet}.
Specifically, a tracklet is defined as a set of robot instantaneous states:}
\begin{multline}
\label{eq:tracklet}
  tr_t(s_t, a_t) = [s^{s}_{t - (u \cdot \Delta t)}, s^{s}_{t - ((u - 1) \cdot \Delta t)}, \ldots, \\ s^{s}_{t}, s^{s}_{t + \Delta t}, \ldots, s^{s}_{t + ((1 + f) \cdot \Delta t)}],
\end{multline}
\zhi{where each state contains the robot's position and velocity, denoted as $s^{s}_t = [p_x, p_y, v_x, v_y]_t$;
$s^{s}_{t - (u \cdot \Delta t)}, s^{s}_{t - ((u - 1) \cdot \Delta t)}, \ldots$ represents $u$ previous states;
$s^{s}_t$ refers to the current state;
$s^{s}_{t + \Delta t}$ represents the state of applying $a_t$ to $s^{s}_t$;
$\ldots, s^{s}_{t + ((1 + f) \cdot \Delta t)}$ represents $f$ future states, which are linearly predicted $f$ times based on $a_t$.
The values of $u$ and $f$ are currently determined by experience, in our experiments $u = 11, f = 3$.}

\zhi{To evaluate the sociality of tracklets we employ a binary classification, i.e. social or non-social.
To this end, a social value function is defined as:
}
\begin{equation}
  SV_t(tr_t(s_t, a_t)) = \psi_{sv} (tr_t(s_t, a_t), W_{sv}),
\end{equation}
where $\psi_{sv}(\cdot)$ is a gated recurrent unit (GRU)~\cite{cho2014learning} with multi-layer perceptron (MLP) and $W_{sv}$ are the weights of it.
The GRU-MLP model is visualized in Fig.~\ref{fig:SVNN_example}, which contains four fully-connected layers, with ReLU nonlinear activation function and batch normalization.

Therefore, the final optimal strategy for robot navigation with the social module is:
\begin{multline}
  \pi^*(s^{jn}_t) = \arg \max_{a_t}[R_t(s^{jn}_{t}, a_{t}) +\\
  + \gamma^{\Delta t\cdot v_{pref}}\cdot V(s^{jn}_{t + \Delta t}) + k_s \cdot SV_t(tr_t(s^{jn}_t, a_t))],
\end{multline}
where $k_s$ is the coefficient of social importance during robot navigation, which is set to 0.6 in our experiments, and
$\gamma^{\Delta t\cdot v_{pref}}\cdot V(s^{jn}_{t + \Delta t}) + k_s \cdot SV_t(tr_t(s^{jn}_t, a_t))$ responds explicitly to our introduction of human factor $\gamma H(s,a)$ in Section~\ref{sec:Problem Formulation}.

\subsection{Online Social Context Learning}

The social performance of robots is challenged by changing social contexts.
Although this problem can theoretically be solved by building massive complete datasets to train general and universal models, in practice, on the one hand, the cost of implementing such idea can be very high, and on the other hand, it is inherently unable to support long-term autonomy of mobile robots because the assumption that the dataset contains all social contexts is invalid~\cite{yz23aaai}.
Therefore, a mechanism is introduced in this letter to update the proposed social model online, triggered by the latter's detection of differences between its internal and external social contexts.
If the difference exceeds a tolerance, the module needs to be retrained online on the external context, and the training is iterated until the difference falls within the tolerance.
It is worth pointing out that considering the general computing power of current mobile robots, updating only the social module instead of the entire system, i.e., the DRL layer and the social module, is an important guarantee for the robot's "fast learning and immediate deployment".

\zhi{The proposed online social context learning is detailed in Algorithm~\ref{alg1}, which is an implementation of $f_d$ in Eq.~\ref{eq.:f_d}.
First, a buffer is set up to continuously collect the states of the robot and the humans around it (line 2).
When the buffer is full (line 10), the states that constitute the robot's last tracklet and the states that constitute people's last tracklets are transferred separately (lines 11-12). The definition of human tracklets is identical to the robot ones in Eq.~\ref{eq:tracklet} but with human positions and velocities. To determine future states, $a_t$ is equal to the current velocity of the human agent.
When the number of stored robot tracklets reaches a threshold (line 13), the efficiency of the social module is analyzed.}

\zhi{Next, the labels of the states in a part of the human last $K_{up}$ tracklets (lines 12 and 16), i.e., social or non-social, need to be determined.
Existing methods are often based on a strong assumption that all human behavior is social~\cite{kretzschmar2016socially, adeli2020socially, vintr20iros}.
Based on our previous findings~\cite{okunevich2023human}, we propose here a more elegant labeling method.
A metric named \emph{extra distance ratio} is used, which is defined as:
\begin{equation}
    R_{dist} = d_s / d_a,
\end{equation}
where $d_s$ represents the euclidean distance from the start point to the end point in a tracklet, while $d_a$ represents the actual length of the tracklet.
The higher the value of $R_{dist}$, the better the sociality.
Therefore, if a tracklet's $R_{dist}$ is higher than a preset threshold, all states on the tracklet are considered social, otherwise non-social (line 14).
Through this method, the robot obtains information about the external social context.
Then, through the output of the social module, it becomes aware of its internal social context (line 15).} 

\zhi{Finally, if the binary accuracy of the two contexts is below a threshold (the lower the accuracy, the greater the difference, line 17), the social model is updated (line 21).
To do so, the aforementioned labeling method is still used to label the human set $Tr_h$ (line 18) and the robot set $Tr_r$ (line 19) respectively.
Then the training set $D_{new}$ consists of $Tr_h$ and non-social samples in $Tr_r$ (line 20).
Additionally, to prevent model overfitting, the robot and human sets are purged (line 22) and old data will not be used in the next iteration.
$L_{trak}$, $K_{up}$ and $K_{acc}$ are hyperparameters and need to be fine tuned.
In our experiments, $L_{trak} = 16$, $K_{up} = 3$ and $K_{acc} = 0.5$.}

\begin{algorithm}[htbp]
\caption{
    Online Social Context Learning}\label{alg1}
    \textbf{Input:} $\psi_{sv}(\cdot, W_{sv})$: the social neural network\\
    $T_t$: the buffered human and robot states at time $t$\\
    $L_{trak}$: the dimension of the input to $\psi_{sv}$\\ 
    $Tr_r$: the set of robot tracklets\\
    $Tr_h$: the set of human tracklets\\
    $Tr^{'}_h$: the set of last human tracklets\\
    $K_{up}$: the update threshold\\
    $K_{acc}$: the accuracy threshold\\
    \textbf{Output:} $\psi_{sv}(\cdot, W_{sv})$: the social neural network\\
    \If {$|T_t|$ \% $L_{trak}$ = $0$}{
        $Tr_r$ $\leftarrow$ The last robot tracklet ($L_{trak} \times s^{s}_r$)\\
        $Tr_h$, $Tr^{'}_h$ $\leftarrow$ The last human tracklets $(L_{trak} \times s^{s}_{h}) \times n$\\
        \If {$|Tr_r|$ \% $K_{up}$ = $0$}{
            $Y_{Tr^{'}_h}$ $\leftarrow$ label$(Tr^{'}_h)$;\\
            $\hat{Y}_{Tr^{'}_h}$ $\leftarrow$
            $\psi_{sv}(Tr^{'}_h, W_{sv})$;\\
            $Tr^{'}_h$ $\leftarrow$ $\emptyset$;\\
            \If {binary\_acc$(Y_{Tr^{'}_h}, \hat{Y}_{Tr^{'}_h}) < K_{acc}$}{
                $Y_{Tr_h}$ $\leftarrow$ label$(Tr_h)$;\\
                $Y_{Tr_r}$ $\leftarrow$ label$(Tr_r)$;\\
                $D_{new} \leftarrow [Tr_h, Y_{Tr_h}] \cup \{x, y \in Tr_r, Y_{Tr_r} \mid y = 0\}$;\\
                $W_{sv}^{new}$ $\leftarrow$ train$(D_{new})$;\\
                $Tr_r$, $Tr_h$ $\leftarrow$ $\emptyset$;\\
            }
        }
    }
\end{algorithm}

\section{Evaluation}

\zhi{In this section we introduce the initialization of the social module, simulation, and real robot experiments.}

\subsection{Social Module Initialization}

\zhi{The social module can be initialized in online learning during deployment~\cite{yz17iros, yz18iros} or offline training using simulators or datasets.
We adopt the latter because of its higher experimental efficiency.
Specifically, the Thor-Magni dataset~\cite{schreiter2024th} is used, which includes five different HRI scenarios with the number of participants ranging from four to nine.
The human trajectories were recorded by a high-precision motion capture system and are used by us as social examples.
We then apply the well-known Optimal Reciprocal Collision Avoidance (ORCA)~\cite{van2011reciprocal} method to generate corresponding robot trajectories according to the human ones, which are labeled as non-social. All trajectories are segmented into 41675 tracklets, each contains 16 points (i.e. $L_{trak}$).
The mean $R_{dist}$ of the social and non-social tracklets are 0.88 [SD = 0.20] and 0.85 [SD = 0.21], respectively.
After shuffling, 70\% of them is used for training and 30\% for validation.
The social neural network is trained in 50 epochs with a batch size of 32.
The optimizer is RMSProp with a learning rate of 0.001.
The criterion is a binary cross-entropy loss, which results in an accuracy of 89.69\%.
For more details on the non-social robot trajectory generation and the social model training, please refer to our open source repository.}

\subsection{Simulation Results}

\zhi{We first evaluate our proposed method using the SARL simulator~\cite{chen2019crowd} widely used by the community~\cite{yang2023st, cheng2024multi}.
The original experiment involved a robot agent moving between two points at a distance of $8~m$, while 5 human agents also moved between two other points at the same distance.
Since the lines connecting a pair of points all intersect at one place, HRI occurs when the robot and human agents move.
The only difference between the test cases is the starting position of the human agents.
SARL demonstrated excellent results under these settings, with a success rate of 100\%.}

\zhi{The original experiments of SARL are enriched in this letter.
First, the number of human agents is increased, ranging from 5 to 10 in steps of 1.
When the number of agents reaches 10, the simulation will start again from 5, and the process repeats.
Second, the test cases are complicated to be more realistic.
The robot agent is asked to reach multiple goals in sequence, rather than one.
The robot has 25 seconds to reach the first goal as per~\cite{chen2019crowd}. After reaching the goal, the robot gets an additional 25 seconds to reach the next goal.
Third, the robot agent needs to traverse distances of $40~m$ and $68~m$, and the human agent is controlled by ORCA (as per~\cite{chen2017decentralized, chen2019crowd, yang2023st}) and SF (first use) respectively, thus constituting 4 experimental categories: \textit{Short-ORCA}, \textit{Short-SF}, \textit{Long-ORCA} and \textit{Long-SF}.}

\zhi{Two preset experimental scenarios\footnote{\url{https://github.com/Nedzhaken/SOCSARL-OL}} are used.
In one, all human agents are randomly initialized within a circle of radius $4~m$, and in the other, the circle is replaced by a square with a side length of $10~m$.
Each scenario contains 250 test cases, with the number of agents evenly distributed.
The radius and preferred speed of the robot and human agents are $0.3~m$ and $1~m/s$ respectively (cf. Eq.~\ref{eq:state}).
We take the heuristic-based ORCA method without sociality as the baseline and conduct comparative experiments with four other learning-based methods, including CADRL~\cite{chen2017decentralized}, LSTM-RL~\cite{everett2018motion}, SARL~\cite{chen2019crowd} and ST$^2$~\cite{yang2023st}.
We compare these methods with our own: SOCSARL and SOCSARL-OL. SOCSARL is the combination of SARL and the social module. SOCSARL-OL is the combination of SARL and the social module plus the online learning mechanism of the social module. We do not include online learning in SOCSARL for ablation purposes.
All methods were trained in the ORCA environment in the SARL simulator. The training scenario was to reach one goal in $8~m$ from the start robot position~\cite{chen2019crowd}.
Three common performance metrics~\cite{chen2017socially, chen2019crowd, yang2023st} including navigation success rate, collision rate, and average navigation time of successful cases are used for evaluation.
Success is the ratio of successful test cases to the total number of test cases. A successful case is when the robot reaches all goals before timeout without collision with human agents. Collision is the ratio of test cases where the robot collides with a human to the total number of test cases.
A collision in the simulator would correspond to a close encounter with a human in the real world. Such an encounter would still result in perceived discomfort with the cost being much lower than an actual collision.
Time is the average time duration of all successful test cases.}
The experimental results are shown in Table~\ref{tabl:T1}.

\begin{table}[t]
  \centering
  \caption{Evaluation Results of Different Methods}
  \label{tabl:T1}
  \begin{tabular}{|l|c|c|c|c|}
    \hline
    \textbf{Method} & \textbf{Success} & \textbf{Collision} & \textbf{\text{Time} (s)} & $\textbf{R}_{\textbf{dist}}$\\
    \hline
    \hline
    \multicolumn{5}{|c|}{\textit{Short-ORCA}}\\
    \hline
    \rowcolor{pink}
    ORCA~\cite{van2011reciprocal} & 0.91 & 0.00 & 53.48 & 0.964\\
    \hline
    CADRL~\cite{chen2017decentralized} & 0.06 & 0.93 & 59.49 & 0.828\\
    \hline
    LSTM-RL~\cite{everett2018motion} & 0.23 & 0.62 & 48.55 & 0.886\\
    \hline
    SARL~\cite{chen2019crowd} & 0.71 & 0.25 & 50.94 & 0.928\\
    \hline
    ST$^2$~\cite{yang2023st} & 0.54 & 0.46 & \textbf{48.50} & 0.899\\
    \hline
    \rowcolor{lightgray}
    SOCSARL (ours) & 0.70 & 0.25 & 50.84 & 0.929\\
    \hline
    \rowcolor{lightgray}
    SOCSARL-OL (ours) & \textbf{0.77} & \textbf{0.23} & 49.17 & \textbf{0.933}\\
    \hline
    \hline
    \multicolumn{5}{|c|}{\textit{Short-SF}}\\
    \hline
    \rowcolor{pink}
    ORCA~\cite{van2011reciprocal} & 0.01 & 0.99 & 44.42 & 0.99\\
    \hline
    CADRL~\cite{chen2017decentralized} & 0.19 & 0.79 & 54.83 & 0.898\\
    \hline
    LSTM-RL~\cite{everett2018motion} & 0.13 & 0.74 & 47.41 & 0.896\\
    \hline
    SARL~\cite{chen2019crowd} & 0.77 & \textbf{0.19} & 48.96 & 0.957\\
    \hline
    ST$^2$~\cite{yang2023st} & 0.60 & 0.40 & \textbf{45.96} & 0.935\\
    \hline
    \rowcolor{lightgray}
    SOCSARL (ours) & 0.77 & \textbf{0.19} & 48.92 & 0.957\\
    \hline
    \rowcolor{lightgray}
    SOCSARL-OL (ours) & \textbf{0.81} & \textbf{0.19} & 47.61 & \textbf{0.958}\\
    \hline
    \hline
    \multicolumn{5}{|c|}{\textit{Long-ORCA}}\\
    \hline
    \rowcolor{pink}
    ORCA~\cite{van2011reciprocal} & 0.84 & 0.00 & 91.29 & 0.960\\
    \hline
    CADRL~\cite{chen2017decentralized} & 0.01 & 0.97 & 105.83 & 0.833\\
    \hline
    LSTM-RL~\cite{everett2018motion} & 0.06 & 0.78 & 88.23 & 0.891\\
    \hline
    SARL~\cite{chen2019crowd} & 0.54 & \textbf{0.38} & 90.3 & 0.921\\
    \hline
    ST$^2$~\cite{yang2023st} & 0.33 & 0.67 & \textbf{81.17} & 0.908\\
    \hline
    \rowcolor{lightgray}
    SOCSARL (ours) & 0.54 & \textbf{0.38} & 90.05 & 0.921\\
    \hline
    \rowcolor{lightgray}
    SOCSARL-OL (ours) & \textbf{0.62} & \textbf{0.38} & 85.87 & \textbf{0.927}\\
    \hline
    \hline
    \multicolumn{5}{|c|}{\textit{Long-SF}}\\
    \hline
    \rowcolor{pink}
    ORCA~\cite{van2011reciprocal} & 0.00 & 1.00 & - & -\\
    \hline
    CADRL~\cite{chen2017decentralized} & 0.09 & 0.88 & 96.86 & 0.898\\
    \hline
    LSTM-RL~\cite{everett2018motion} & 0.01 & 0.84 & 83.29 & 0.894\\
    \hline
    SARL~\cite{chen2019crowd} & 0.56 & 0.36 & 85.89 & 0.956\\
    \hline
    ST$^2$~\cite{yang2023st} & 0.42 & 0.58 & \textbf{77.03} & 0.943\\
    \hline
    \rowcolor{lightgray}
    SOCSARL (ours) & 0.57 & \textbf{0.35} & 86.00 & 0.956\\
    \hline
    \rowcolor{lightgray}
    SOCSARL-OL (ours) & \textbf{0.64} & \textbf{0.35} & 82.66 & \textbf{0.959}\\
    \hline
  \end{tabular}
\end{table}

\zhi{First, the results of the baseline method ORCA confirm the necessity of introducing SF into our experiments.
Because when both the robot and human agents use ORCA, the former shows performance that exceeds all other tested methods.
When the human agents are controlled by SF, the ORCA-based robot agent shows the worst performance among all methods.
These results illustrate the importance of consistency of the robot's internal and external social contexts.
From a sociological perspective, sociality is a consensus that a social agent may not function well in a non-social environment, and vice versa.
The performance differences of other methods in the ORCA and SF categories also support this claim.
Specifically, the social methods SARL, ST$^2$, and ours perform better in the SF-based social environments than in the ORCA-based non-social environments.
While the non-social method LSTM-RL performs better in the ORCA-based environments.
Although CADRL is also a non-social method, it is sensitive to the distance between agents and thus coincidentally closer to the social context, thus performing better in the SF-based environments.}

\zhi{Second, the enriched experimental settings pose challenges to the SOTA methods, but the overall performance ranking of SARL, LSTM-RL, and CADRL is consistent with that reported in~\cite{chen2019crowd}.
Specifically, SARL shows the best results among the three, recalling that SARL is the development of LSTM-RL, which is the development of CADRL.
The overall performance of ST$^2$ is inferior to that of SARL, which is different from~\cite{yang2023st}.
This can be interpreted as SARL having an advantage in dealing with complex environments.}

\zhi{Third, it can be seen that our proposed method outperforms other learning-based ones overall.
The $R_{dist}$ metric demonstrates the effectiveness of our improvement to the original SARL reward function (cf. Eq.~\ref{eq:reward}).
ST$^2$ shows fast times in four categories but at the expense of performance in the other two metrics.
In addition, it can be seen that the performance of our method without the OL module is comparable to that of SARL.
This shows that, on the one hand, the proposed social module basically recognizes the sociality of SARL output without online updates.
On the other hand, the improvement of the robot's social performance is indeed achieved by updating the social module online.
}

\zhi{Finally, we provide insights into the robustness of different methods in different environments.
According to the two metrics of ``success rate'' and ``collision rate'', our SOCSARL-OL has a maximum difference of 4\% in different category pairs (i.e. Short-ORCA and Short-SF, Long-ORCA and Long-SF), ST$^2$ has 9\%, SARL has 6\%, LSTM-RL has 10\%, and CADRL has 13\%.
Our method shows the best robustness.
This again demonstrates the superiority of online learning, while SOCSARL without online learning has a difference of up to 7\%.}

The limitation of our method is that it does not use non-human obstacle information. As our evaluation focused on comparing methods that do not use this information (SARL, ST2, etc.), we did not consider this problem. The non-human obstacles or maps of the environment can be a source of important information for understanding human behavior.
\new{Surrounding the robot with human agents often causes collisions. Future work could optimize SARL training specifically for high-density dynamic crowds.}

\subsection{Real Robot Experiments}

\begin{figure}[t]
  \centering
  \subfigure[The example of non-social ORCA (left column) and social SOCSARL-OL behavior (right column). The top images are from the simulation and the bottom images are from the real robot experiment. The human social zones, which are crossed by the robot are red, other human social zones are green. The blue arrow is the robot's direction of movement.] {
    \includegraphics[width=0.9\columnwidth]{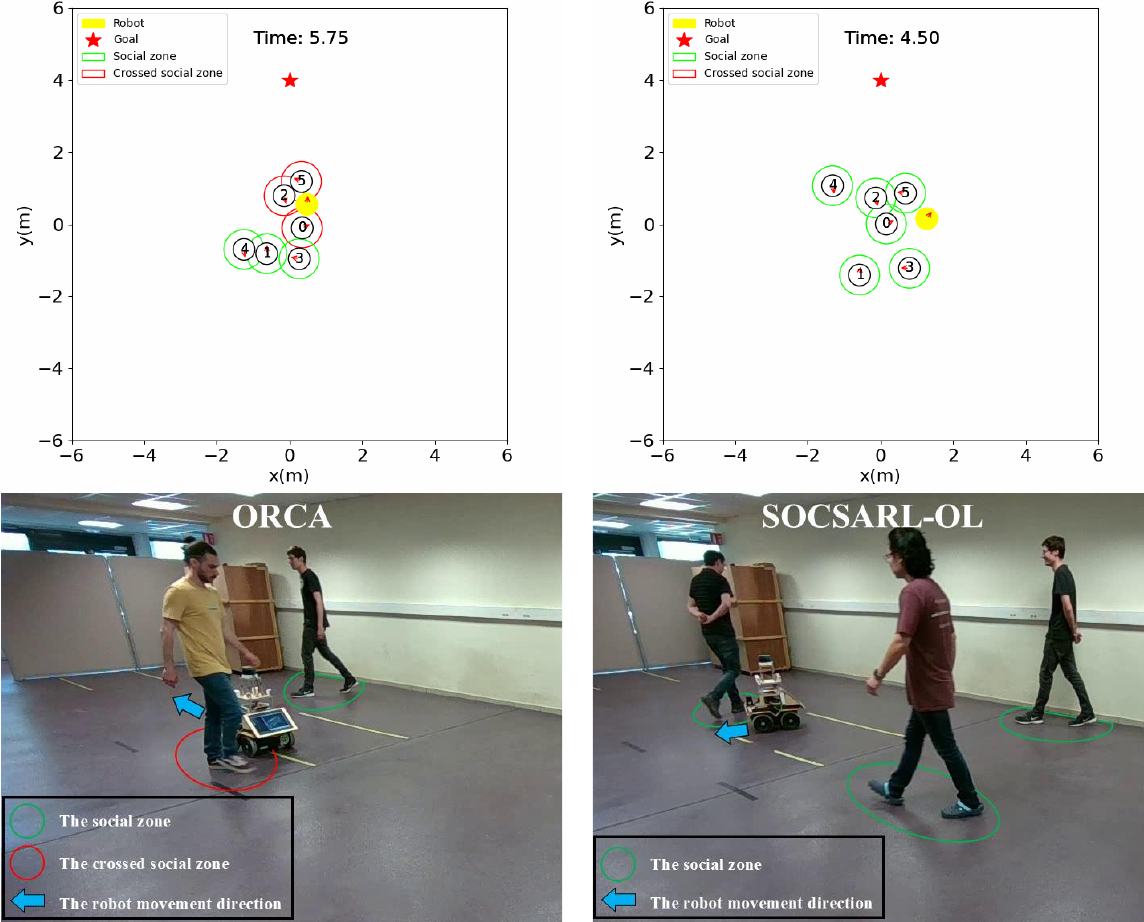}
    \label{fig:experiment_real}
  }
  \subfigure[Mean and standard deviation of the scores given by twelve human participants in the real robot experiment.] {
    \includegraphics[width=0.9\columnwidth]{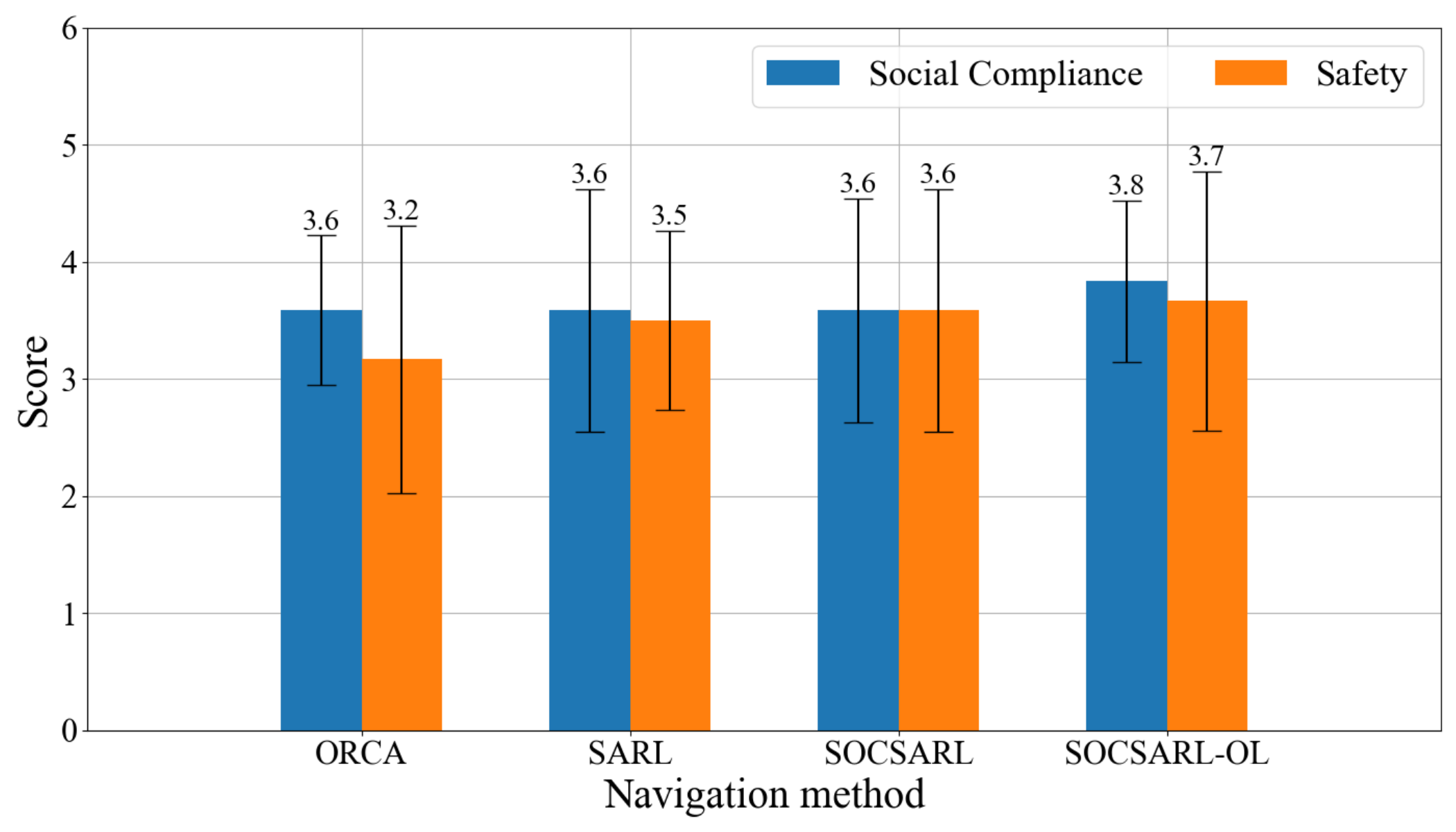}
    \label{fig:experiment_real_plot}
  }
  \caption{Real robot experiment and its results.}
  \label{fig: experiment}
\end{figure}

\zhi{Our method is evaluated using a real robot we developed~\cite{okunevich2023human, okunevich2024open}.}
We wanted to demonstrate the effectiveness of the robot social navigation system and measure how humans felt about interacting with the robot.
We replicated the experimental scenario from the SARL paper~\cite{chen2019crowd}, as shown in Fig.~\ref{fig:experiment_real}.
People randomly walked around the room for 30 seconds, while the robot had to cross the room.
The human tracking system from our previous work was applied to obtain human tracklets~\cite{okunevich2024open}. In our work, we did not aim to prevent the robot's movements from distracting the human group. However, overlapping social zones of people standing close together force the robot to move around without disturbing the group.
12 human participants were divided into 3 groups of 5/5/2 for the experiment.

ORCA, SARL, SOCSARL, and SOCSARL-OL were evaluated during the experiment.
Methods were presented randomly.
The real-time performance of ORCA, SARL, SOCSARL, and SOCSARL-OL are 125, 5.5, 4.9, and 4.5 $Hz$ respectively.
After the experiment, the participants answered the following two questions for each method as per~\cite{karnan2022socially}:
1) On a scale of 1 to 5, how ``socially compliant'' do you think the robot was (think of social compliance as how considerate the robot was of your presence)?
2) On a scale of 1 to 5, how ``safe'' did you feel around the robot?

The results of the survey are shown in Fig.~\ref{fig:experiment_real_plot}.
It can be seen that SARL, SOCSARL, and SOCSARL-OL make people feel safer than ORCA. The social feeling of all methods are comparable.
Later, we conducted a short interview with the participants.
Their main opinion was that they were confused by the avoidance action of SARL, SOCSARL, and SOCSARL-OL methods.
This was attributed to the social distance condition in the reward function (cf. Eq.~\ref{eq:reward}), which made the robot start the avoidance action very early.
This was inconsistent with the participants' stereotypes of mobile robots.
People generally believed that the robot could get closer to them before reacting.
However, this was contrary to the safety requirement of the robot's movement.

\section{Conclusions}

\new{This letter presented an online context learning method for socially compliant robot navigation, which combines DRL and ORL to help robots adapt to changing social contexts and improve its social performance.}
The experimental results demonstrated that our method surpasses the SOTA ones.
Future work includes further evaluation of the real-world performance of online context learning by running the robot over longer periods of time in larger spaces, and assessing the robustness of the system in different real-world public spaces.
\new{Moreover, the experiment would include both holonomic and non-holonomic robots for a more comprehensive evaluation, and applying non-human obstacle information in the navigation process or improving the structure of the social module are also desired improvements in the future.}

\section*{Acknowledgments}

We thank RoboSense for sponsoring the 3D lidar, UTBM CRUNCH Lab for support in robotic instrumentation, and all those involved in the experiments.


\bibliographystyle{IEEEtran} 
\bibliography{references}

\begin{thebibliography}{10}
\providecommand{\url}[1]{#1}
\csname url@samestyle\endcsname
\providecommand{\newblock}{\relax}
\providecommand{\bibinfo}[2]{#2}
\providecommand{\BIBentrySTDinterwordspacing}{\spaceskip=0pt\relax}
\providecommand{\BIBentryALTinterwordstretchfactor}{4}
\providecommand{\BIBentryALTinterwordspacing}{\spaceskip=\fontdimen2\font plus
\BIBentryALTinterwordstretchfactor\fontdimen3\font minus
  \fontdimen4\font\relax}
\providecommand{\BIBforeignlanguage}[2]{{%
\expandafter\ifx\csname l@#1\endcsname\relax
\typeout{** WARNING: IEEEtran.bst: No hyphenation pattern has been}%
\typeout{** loaded for the language `#1'. Using the pattern for}%
\typeout{** the default language instead.}%
\else
\language=\csname l@#1\endcsname
\fi
#2}}
\providecommand{\BIBdecl}{\relax}
\BIBdecl

\bibitem{zhimon20jist}
Z.~Yan, S.~Schreiberhuber, G.~Halmetschlager, T.~Duckett, M.~Vincze, and
  N.~Bellotto, ``Robot perception of static and dynamic objects with an
  autonomous floor scrubber,'' \emph{Intelligent Service Robotics}, vol.~13,
  no.~3, pp. 403--417, 2020.

\bibitem{perminov2021ultrabot}
S.~Perminov, N.~Mikhailovskiy, A.~Sedunin, I.~Okunevich, I.~Kalinov,
  M.~Kurenkov, and D.~Tsetserukou, ``Ultrabot: Autonomous mobile robot for
  indoor uv-c disinfection,'' in \emph{Proceedings of CASE}, 2021, pp.
  2147--2152.

\bibitem{vintr20iros}
T.~Vintr, Z.~Yan, K.~Eyisoy, F.~Kubis, J.~Blaha, J.~Ulrich, C.~S. Swaminathan,
  S.~M. Mellado, T.~Kucner, M.~Magnusson, G.~Cielniak, J.~Faigl, T.~Duckett,
  A.~J. Lilienthal, and T.~Krajn{\'{\i}}k, ``Natural criteria for comparison of
  pedestrian flow forecasting models,'' in \emph{Proceedings of IROS}, 2020,
  pp. 11\,197--11\,204.

\bibitem{okunevich2023human}
I.~Okunevich, V.~Hilaire, S.~Galland, O.~Lamotte, L.~Shilova, Y.~Ruichek, and
  Z.~Yan, ``Human-centered benchmarking for socially-compliant robot
  navigation,'' in \emph{Proceedings of ECMR}, 2023, pp. 1--7.

\bibitem{shahrezaie2022advancing}
R.~S. Shahrezaie, B.~N. Manalo, A.~G. Brantley, C.~R. Lynch, and
  D.~Feil-Seifer, ``Advancing socially-aware navigation for public spaces,'' in
  \emph{Proceedings of RO-MAN}, 2022, pp. 1015--1022.

\bibitem{benedictis2023dichotomic}
R.~D. Benedictis, A.~Umbrico, F.~Fracasso, G.~Cortellessa, A.~Orlandini, and
  A.~Cesta, ``A dichotomic approach to adaptive interaction for socially
  assistive robots,'' \emph{User Modeling and User-Adapted Interaction},
  vol.~33, no.~2, pp. 293--331, 2023.

\bibitem{chen2017socially}
Y.~F. Chen, M.~Everett, M.~Liu, and J.~P. How, ``Socially aware motion planning
  with deep reinforcement learning,'' in \emph{Proceedings of IROS}, 2017, pp.
  1343--1350.

\bibitem{chen2019crowd}
C.~Chen, Y.~Liu, S.~Kreiss, and A.~Alahi, ``Crowd-robot interaction:
  Crowd-aware robot navigation with attention-based deep reinforcement
  learning,'' in \emph{Proceedings of ICRA}, 2019, pp. 6015--6022.

\bibitem{yang2023st}
Y.~Yang, J.~Jiang, J.~Zhang, J.~Huang, and M.~Gao, ``St$^{2} $:
  Spatial-temporal state transformer for crowd-aware autonomous navigation,''
  \emph{RA-L}, vol.~8, no.~2, pp. 912--919, 2023.

\bibitem{schreiter2024th}
T.~Schreiter, T.~R. de~Almeida, Y.~Zhu, E.~G. Maestro, L.~Morillo-Mendez,
  A.~Rudenko, L.~Palmieri, T.~P. Kucner, M.~Magnusson, and A.~J. Lilienthal,
  ``Thor-magni: A large-scale indoor motion capture recording of human movement
  and robot interaction,'' \emph{arXiv preprint arXiv:2403.09285}, 2024.

\bibitem{karnan2022socially}
H.~Karnan, A.~Nair, X.~Xiao, G.~Warnell, S.~Pirk, A.~Toshev, J.~Hart,
  J.~Biswas, and P.~Stone, ``Socially compliant navigation dataset (scand): A
  large-scale dataset of demonstrations for social navigation,'' \emph{RA-L},
  vol.~7, no.~4, pp. 11\,807--11\,814, 2022.

\bibitem{tai2018socially}
L.~Tai, J.~Zhang, M.~Liu, and W.~Burgard, ``Socially compliant navigation
  through raw depth inputs with generative adversarial imitation learning,'' in
  \emph{Proceedings of ICRA}, 2018, pp. 1111--1117.

\bibitem{bhattacharyya2019simulating}
R.~P. Bhattacharyya, D.~J. Phillips, C.~Liu, J.~K. Gupta, K.~Driggs-Campbell,
  and M.~J. Kochenderfer, ``Simulating emergent properties of human driving
  behavior using multi-agent reward augmented imitation learning,'' in
  \emph{Proceedings of ICRA}, 2019, pp. 789--795.

\bibitem{chen2017decentralized}
Y.~F. Chen, M.~Liu, M.~Everett, and J.~P. How, ``Decentralized
  non-communicating multiagent collision avoidance with deep reinforcement
  learning,'' in \emph{Proceedings of ICRA}, 2017, pp. 285--292.

\bibitem{everett2018motion}
M.~Everett, Y.~F. Chen, and J.~P. How, ``Motion planning among dynamic,
  decision-making agents with deep reinforcement learning,'' in
  \emph{Proceedings of IROS}, 2018, pp. 3052--3059.

\bibitem{liu2021decentralized}
S.~Liu, P.~Chang, W.~Liang, N.~Chakraborty, and K.~Driggs-Campbell,
  ``Decentralized structural-rnn for robot crowd navigation with deep
  reinforcement learning,'' in \emph{Proceedings of ICRA}, 2021, pp.
  3517--3524.

\bibitem{singamaneni2021human}
P.~T. Singamaneni, A.~Favier, and R.~Alami, ``Human-aware navigation planner
  for diverse human-robot interaction contexts,'' in \emph{Proceedings of
  IROS}, 2021, pp. 5817--5824.

\bibitem{liu2021lifelong}
B.~Liu, X.~Xiao, and P.~Stone, ``A lifelong learning approach to mobile robot
  navigation,'' \emph{RA-L}, vol.~6, no.~2, pp. 1090--1096, 2021.

\bibitem{yz23aaai}
Z.~Yan, L.~Sun, T.~Krajnik, T.~Duckett, and N.~Bellotto, ``Towards long-term
  autonomy: A perspective from robot learning,'' in \emph{AAAI-23 Bridge
  Program on AI \& Robotics}, 2023.

\bibitem{okunevich2024open}
I.~Okunevich, V.~Hilaire, S.~Galland, O.~Lamotte, Y.~Ruichek, and Z.~Yan, ``An
  open-source software-hardware integration scheme for embodied human
  perception in service robotics,'' in \emph{Proceedings of ARSO}, 2024.

\bibitem{yz19auro}
Z.~Yan, T.~Duckett, and N.~Bellotto, ``Online learning for 3d lidar-based human
  detection: experimental analysis of point cloud clustering and classification
  methods,'' \emph{Autonomous Robots}, vol.~44, no.~2, pp. 147--164, 2020.

\bibitem{serhan17tcsvt}
S.~Cosar, G.~Donatiello, V.~Bogorny, C.~G{\'{a}}rate, L.~O. Alvares, and
  F.~Br{\'{e}}mond, ``Toward abnormal trajectory and event detection in video
  surveillance,'' \emph{{IEEE} Trans. Circuits Syst. Video Technol.}, vol.~27,
  no.~3, pp. 683--695, 2017.

\bibitem{cho2014learning}
K.~Cho, B.~Van~Merri{\"e}nboer, C.~Gulcehre, D.~Bahdanau, F.~Bougares,
  H.~Schwenk, and Y.~Bengio, ``Learning phrase representations using rnn
  encoder-decoder for statistical machine translation,'' \emph{arXiv preprint
  arXiv:1406.1078}, 2014.

\bibitem{kretzschmar2016socially}
H.~Kretzschmar, M.~Spies, C.~Sprunk, and W.~Burgard, ``Socially compliant
  mobile robot navigation via inverse reinforcement learning,'' \emph{The
  International Journal of Robotics Research}, vol.~35, no.~11, pp. 1289--1307,
  2016.

\bibitem{adeli2020socially}
V.~Adeli, E.~Adeli, I.~Reid, J.~C. Niebles, and H.~Rezatofighi, ``Socially and
  contextually aware human motion and pose forecasting,'' \emph{RA-L}, vol.~5,
  no.~4, pp. 6033--6040, 2020.

\bibitem{yz17iros}
Z.~Yan, T.~Duckett, and N.~Bellotto, ``Online learning for human classification
  in {3D LiDAR-based} tracking,'' in \emph{Proceedings of IROS}, 2017, pp.
  864--871.

\bibitem{yz18iros}
Z.~Yan, L.~Sun, T.~Duckett, and N.~Bellotto, ``Multisensor online transfer
  learning for 3d lidar-based human detection with a mobile robot,'' in
  \emph{Proceedings of IROS}, 2018, pp. 7635--7640.

\bibitem{van2011reciprocal}
J.~Van Den~Berg, S.~J. Guy, M.~Lin, and D.~Manocha, ``Reciprocal n-body
  collision avoidance,'' in \emph{Proceedings of ISRR}, 2011, pp. 3--19.

\bibitem{cheng2024multi}
C.-L. Cheng, C.-C. Hsu, S.~Saeedvand, and J.-H. Jo, ``Multi-objective
  crowd-aware robot navigation system using deep reinforcement learning,''
  \emph{Applied Soft Computing}, vol. 151, p. 111154, 2024.

\end{thebibliography}

\end{document}